\pdfoutput=1
\documentclass[11pt]{article}

\usepackage[review]{acl}

\usepackage{times}
\usepackage{latexsym}
\usepackage{amsmath}
\usepackage{multirow}
\usepackage{booktabs}
\usepackage{graphicx}

\pdfoutput=1

\usepackage[T1]{fontenc}
\usepackage[utf8]{inputenc}

\usepackage{microtype}

\renewcommand{\paragraph}{\textbf}

\title{Effective General-Domain Data Inclusion for the Machine Translation Task by Vanilla Transformers}

\author{Hassan Soliman \\
  ID: 2576774 \\
  Saarland University \\
  \texttt{s8hasoli@stud.uni-saarland.de} \\}

\begin{document}
\maketitle

\begin{abstract}

One of the vital breakthroughs in the history of machine translation is the development of the Transformer model. Not only it is revolutionary for various translation tasks, but also for a majority of other NLP tasks. In this paper, we aim at a Transformer-based system that is able to translate a source sentence in German to its counterpart target sentence in English. We perform the experiments on the news commentary German-English parallel sentences from the WMT'13 dataset. In addition, we investigate the effect of the inclusion of additional general-domain data in training from the IWSLT'16 dataset to improve the Transformer model performance. We find that including the IWSLT'16 dataset in training helps achieve a gain of 2 BLEU score points on the test set of the WMT'13 dataset. Qualitative analysis is introduced to analyze how the usage of general-domain data helps improve the quality of the produced translation sentences.

\end{abstract}
\section{Introduction}

Machine translation (MT) is a technique that uses machine learning to translate from one language to another. Machine translation has evolved greatly in recent years, allowing users to understand text in a foreign language without any knowledge of the language itself \cite{QiuMM21}.

A classical statistical machine translation (SMT) system was initially employed for the MT task, which relies on separate lossy components, such as word aligners, translation rule extractors, and other features extractors \cite{DabreCK20}. There was no robust end-to-end model for the translation task.

MT using neural methods, referred to as neural machine translation (NMT), has become the dominant paradigm for academic and commercial research \cite{DabreCK20}. This paradigm has proven very successful at learning features from data, resulting in a remarkable breakthrough in the field of MT \cite{he2021attention}. Its success is largely attributed to the use of distributed representations of language that enable end-to-end training of a MT system. The dominant NMT approach in the last few years is Embed - Encode - Attend - Decode.

The state-of-the-art NMT models are almost entirely based on the attention mechanism. Attention models make it easier to relate input sequence units regardless of their spatial or temporal distance. Additionally, attention models make it easier to parallelize sequence data processing. \cite{he2021attention}.

The Transformer model, proposed by \cite{vaswani_attention_2017}, represents an important contribution to natural language processing (NLP) in general and to NMT in particular. It is classified as a branch of sequential neural networks, offering a faster processing speed than recurrent neural networks (RNN) \cite{KatharopoulosV020}. The Transformer is an attention-based model. There are no recurrent or convolutional layers. Because of this, it has a very straightforward and nearly linear structure, which explains its rapid processing speed \cite{QiuMM21}.

The main contributions of this paper include the following: (1) Investigating and carefully describing the different components constituting the Transformer model, (2) Testing the Transformer model with the popular German-English dataset from the 2013 Workshop on Statistical Machine Translation Task (WMT’13) dataset \cite{WMT_2013}, and (3) Proposing the inclusion of a general-domain dataset from the 2016 International Workshop of Spoken Language Translation (IWSLT'16) \cite{IWSLT_2016} in the training dataset.

The rest of the paper is organized as follows: Section 2 presents the literature review, Section 3 describes the Transformer model in detail, Section 4 illustrates the dataset collection and preprocessing, Section 5 describes the experimental setup and evaluation results, and finally, Section 6 provides a summary and conclusion including possible future work.

\section{Literature Review}

MT follows the same pattern as other NLP fields in which deep learning became the predominant approach, called NMT \cite{DabreCK20}. In modern-day research, Transformer-based models are almost mostly used in MT systems \cite{QiuMM21}. They show excellent performance for many language pairs and vastly improve the performance of NMT models \cite{he2021attention}. These models are being used for various MT tasks including the well-known WMT and IWSLT tasks.

In this paper, we aim at a neural approach based on the vanilla Transformer model by \cite{vaswani_attention_2017}. However, there are numerous attempts to enhance the basic Transformer encoder-decoder model architecture as stated by \cite{DabreCK20}. Related to this, \cite{GargPNP19} present an approach to train the Transformer model to produce both accurate translations and word alignments. Another work by \cite{GordonD20} explores the usage of general-domain data together with in-domain data for domain adaptation of the Transformer model. \cite{XiaHTTHQ19} propose to share the weights and model parts of both the encoder and decoder of the Transformer model. Moreover, \cite{abrishami2020machine} utilize a weighted combination of layers' primary input and output of the previous layers in the multi-layer encoder and decoder of the Transformer model. \cite{liu_deep_transformer} show that it is feasible to build standard Transformer-based models with up to 60 encoder layers, achieving new state-of-the-art benchmark results on the WMT English-German translation task. \cite{AraabiM20} show that the effectiveness of the Transformer model under low-resource conditions is highly dependent on the hyperparameter settings. \cite{BaharWAGGMH20} make use of synthetic data to fine-tune the encoder and decoder models for domain adaptation. \cite{QiuMM21} propose an enhanced Transformer model for fast streaming translation methods. In the context of obtaining low latencies, \cite{PhamSNHNASW20} employ the Transformer model but implement relative attention in the attention blocks, following the work by \cite{DaiYYCLS19}, which takes into account the relative distances between words instead of using their absolute position vectors.

In inspiration to some of these works, we adapt the inclusion and usage of a general-domain dataset from IWSLT to improve the Transformer model performance on the WMT dataset.
\section{Transformer Model}
%\section{Methods}
%In this section, we detail the Transformer model by \cite{vaswani_attention_2017} and its attention mechanism.

\subsection{Attention Mechanism}
The attention mechanism deployed by \cite{vaswani_attention_2017} is at the core of the Transformer model, which relies on a few basic vector operations. It is used to process a sentence as an input to construct its contextualized embeddings as an output that are a better representation of its true meaning and the surrounding context. This mechanism is essentially a scaled dot-product self-attention, as shown in Figure \ref{fig:attention_1} by \cite{futrzynski_self_attention}. It can be broken down into the following:

\begin{figure}
    \centering
    \includegraphics[width=.45\textwidth]{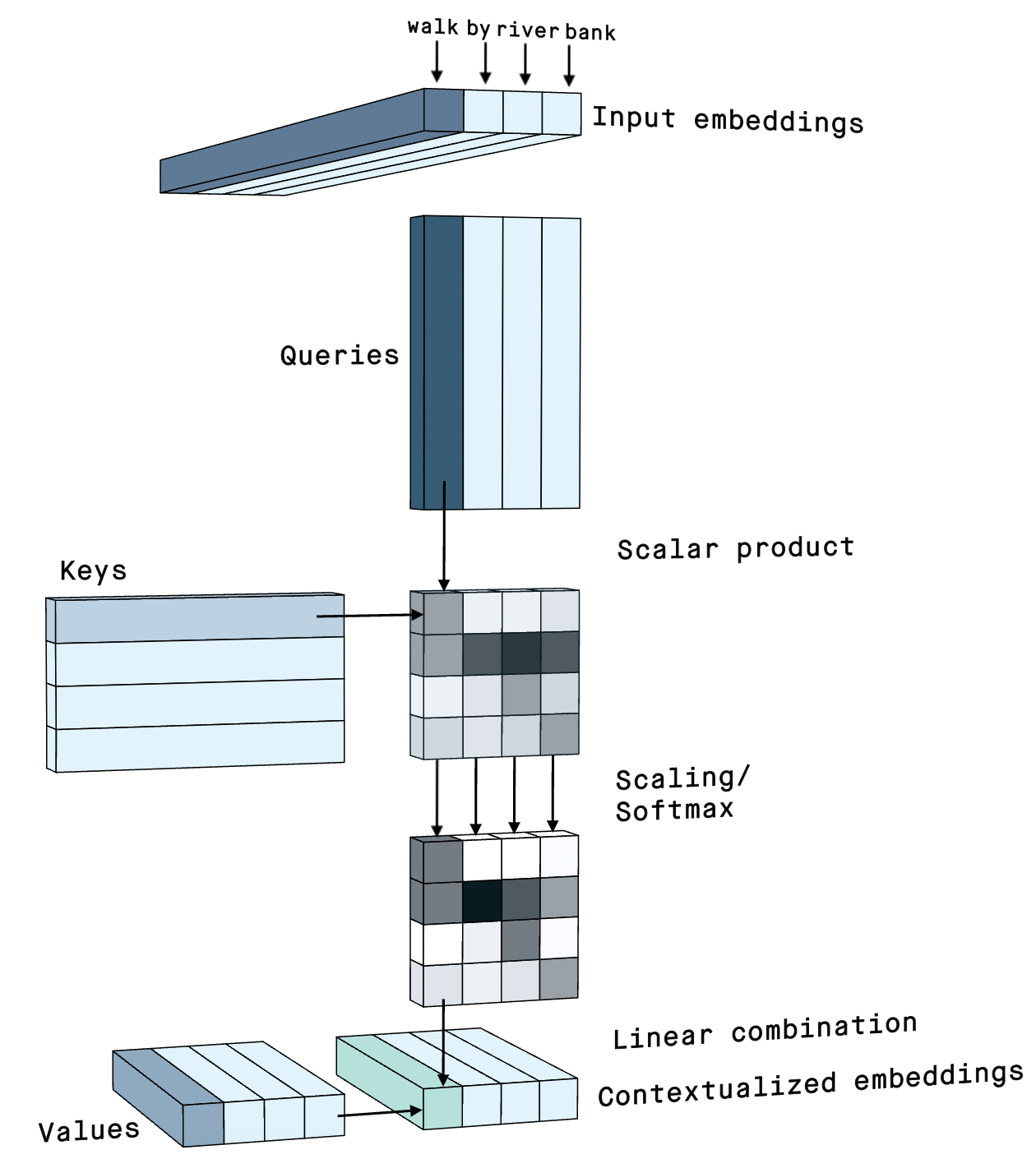}
    \caption{Scaled dot-product self-attention mechanism.}
    \label{fig:attention_1}
\end{figure}

\paragraph{Keys, Queries, and Values.}
The input sentence tokens are converted into embeddings of size 512. Using learned positional encodings, which have the same embedding size of 512, information about the absolute positions of the tokens are included. The two embeddings can then be summed up. For modeling complex relationships between token embeddings, they are represented as keys, queries, and values. Each key, query, and value has its own linear projection layer. Through these projections, we can focus on certain aspects of the embeddings that matter the most in the relationship.

\paragraph{Softmax Activations.}
The scalar dot product between the projections of keys $K$ and queries $Q$ is calculated as shown in Equation \ref{eq:attention}. The dot product is typically scaled-down for numerical stability, where $\sqrt{d_k}$ represents the embedding size of the keys, which in our case is 512. It is then passed through a softmax activation function. This allows non-linear transformations to be modelled.

\begin{equation}
\footnotesize
\begin{split}
    Attention(Q, K, V) = softmax(\frac{QK^T}{\sqrt{d_k}})V
\end{split}
\label{eq:attention}
\end{equation}

The projections of values $V$ corresponding to every input token are taken in proportions according to the results of the softmax function. This results in new contextualized embeddings which are more representative of the input tokens. 

Because the query of the river token is strongly related to the key of the bank token for the term (river bank), the value of the bank token contributes largely to the contextualized embedding of the river token. The final output embeddings are now dependent on the surrounding context.

\begin{figure}
    \centering
    \includegraphics[width=.45\textwidth]{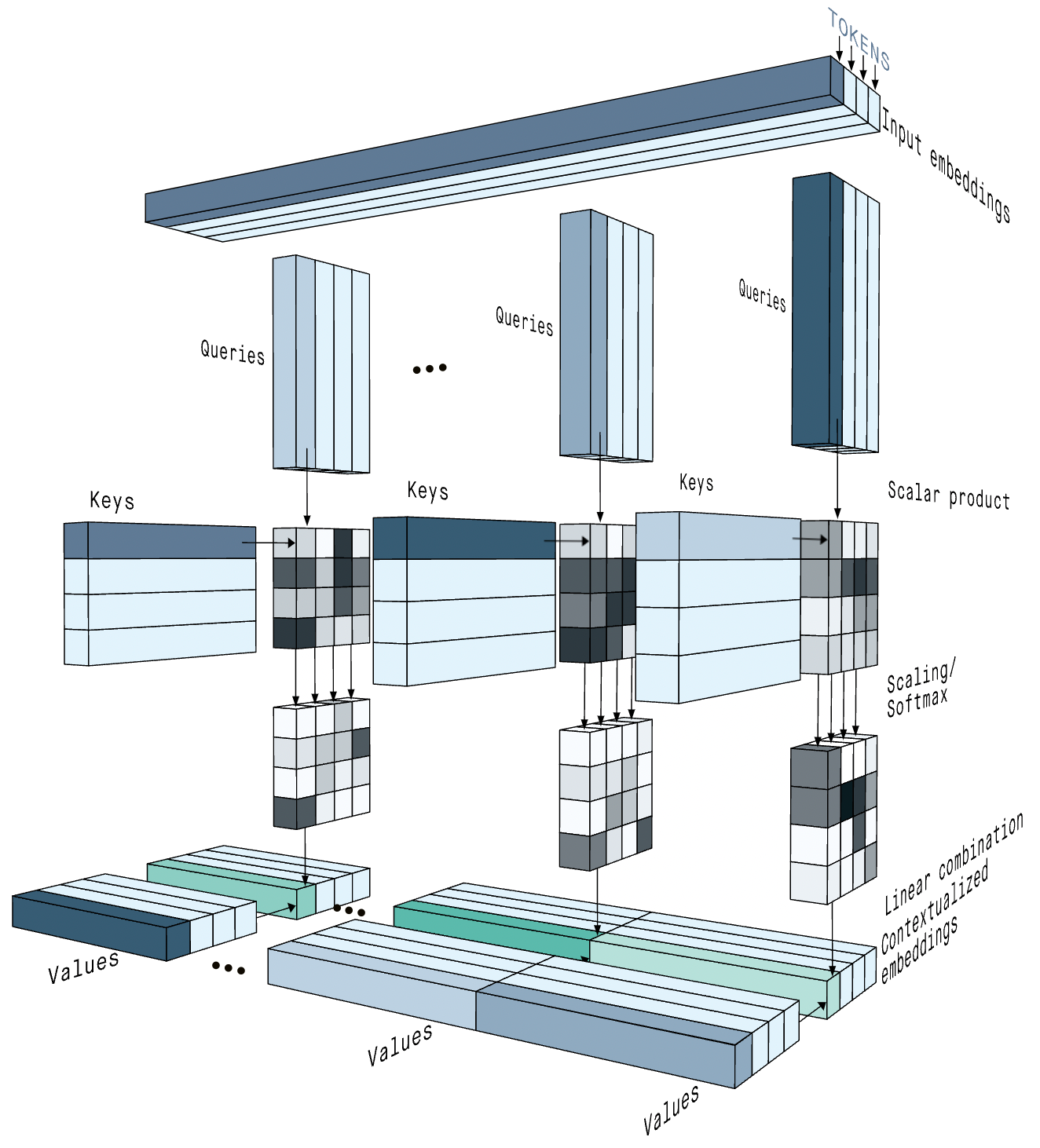}
    \caption{Multi-head attention.}
    \label{fig:attention_2}
\end{figure}

\paragraph{Multi-Head Attention.}
As shown in Figure \ref{fig:attention_2} by \cite{futrzynski_self_attention}, multi-head attention means that the projections of keys, queries, and values are split into heads, which in our case are 8, resulting in projection embeddings of size 64. Each of these projections (heads) focuses on calculating different types of relationships between the tokens and creating the corresponding contextualized embeddings. The contextualized embeddings from each attention head are concatenated to form the final output of the multi-head attention layer. Typically, multiple layers of multi-head attention are used in practice to achieve the best results.

\begin{figure}
    \centering
    \includegraphics[width=.45\textwidth]{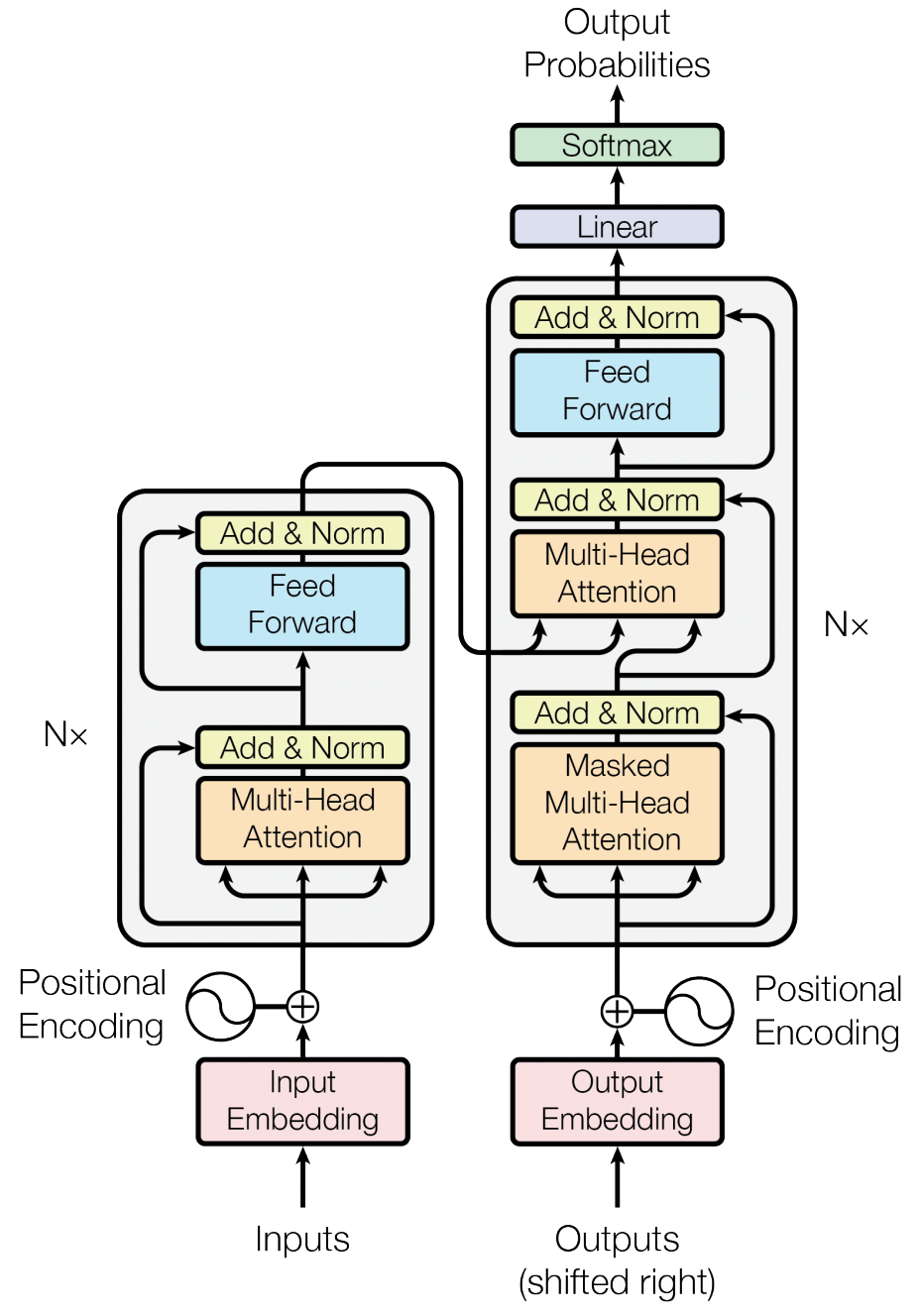}
    \caption{Transformer architecture.}
    \label{fig:transformer}
\end{figure}

\subsection{Encoder}
An encoder consists of a stack of identical layers, which in our case are three. Figure \ref{fig:transformer} shows the structure of one encoder layer. Each layer contains two sub-layers. The input is passed first into a multi-head self-attention layer with a padded mask. This mask ensures that the attention mechanism ignores the token used for padding.

Next, the output is passed through a position-wise fully connected feed-forward layer. Each of these sub-layers is followed by a dropout layer, and a residual connection is added before going through layer normalization.

\subsection{Decoder}
\label{sec:transfer}
The decoder is composed of a stack of identical layers, which in our case are three as well. Figure \ref{fig:transformer} shows the structure of a single decoder layer. Its structure is similar to that of the encoder. In addition to the two sub-layers in the encoder layer, the decoder further adds a third sub-layer in between, which performs multi-head attention over the encoder's output.

First, the input is processed by a multi-head attention layer with a look-ahead mask that prevents the decoder from attending to subsequent positions. The output is then passed through the next sub-layer, which performs multihead attention on the encoder output. The output is then passed through a position-wise fully connected feed-forward layer.

As with the encoder, dropout is used and residual connections are employed around each of the sub-layers, followed by layer normalization. Note that the output embeddings are offset by one position. It ensures that by using the look-ahead mask, the predictions for time step (position) $t$ can only be based on the known outputs at time steps less than $t$.

Finally, to obtain the output probabilities across the vocabulary, the output of the decoder stack is passed through a linear layer and a softmax activation function.

\subsection{Translation Process}
\label{sec:transfer}

In order to translate a sentence from a source language to a target language, the input tokens pass through an embedding layer before being added to their corresponding absolute positional embedding. The sum goes through the encoder stack layers, each of these layers contains a multi-head attention layer and a position-wise fully connected feed-forward network layer.

In the meantime, the 3-layer decoder stack receives as input the translated sequences in the target language. These input tokens are passed through the embedding and positional encoding layers which are added together. They pass then through the first multi-head attention layer. 

This layer's outputs, along with the encoder's outputs, will be passed along to a second multi-head attention layer inside the decoder layer, which is followed by a position-wise fully connected feed-forward layer. After the output passes through the 3-layer decoder stack, it will pass through a feed-forward linear layer and a softmax activation function to determine the output probabilities. The token corresponds to the maximum probability in the softmax layer is produced as an output in each time step to generate the final translated sentence.
\section{Datasets}
%\section{Methods}
%In this section, we detail the entity linking model as well as our transfer method.

\subsection{WMT'13}
The German-English dataset used in this project is from the 2013 Workshop of Statistical Machine Translation (WMT’13) datasets \cite{WMT_2013}. These datasets come from the workshop organizers and are used for the competition on MT tasks. Among these datasets, the German-English news commentary dataset is used in our work. There are also updated versions from WMT after 2013, but the WMT’13 datasets are generally used by many other works in the literature. 

The news commentary dataset originally consists of about 178,000 sentence pairs that contain the same sentence in both German and English. The exact statistics of the dataset can be found in Table \ref{tab:WMT13_statistics}.

\begin{table}[t]
\footnotesize
	\centering\begin{tabular}{@{}l|ll@{}}
    \toprule
                                & \multicolumn{2}{c}{Dataset} \\ \midrule
    Language                    & German       & English      \\ \midrule
    No. of Samples              & 178,793      & 179,011      \\
    No. of Samples Preprocessed & 178,221      & 178,221      \\
    No. of Samples Cleaned      & 177,145      & 177,145      \\
    No. of Samples Unique       & 176,742      & 176,742      \\ \bottomrule
    \end{tabular}
	\caption{WMT'13 dataset statistics and processing.}
	\label{tab:WMT13_statistics}
\end{table}

In our experiments, we use the following data processing steps: (i) data is preprocessed using Moses data preprocessing tool \cite{Koehn_Moses_07}, which normalizes punctuation marks and does word segmentation, (ii) data is cleaned by removing empty sentences, deleting sentences that are obviously not aligned, and controlling long sentences to have a sequence length only from 1 to 80, and (iii) finally, duplicate sentences are removed to obtain unique sentences.
Afterward, the data is split into train, validation, and test sets. The statistics of the splits can be found in Table \ref{tab:WMT13_splits}.

\begin{table}[t]
\footnotesize
	\centering\begin{tabular}{@{}l|ccc@{}}
    \toprule
                   & \multicolumn{3}{c}{Dataset}                                    \\ \midrule
    Splits         & Train                       & Valid & \multicolumn{1}{c}{Test} \\ \midrule
    No. of Samples & \multicolumn{1}{l}{174,742} & 1,000 & 1,000                    \\ \bottomrule
    \end{tabular}
	\caption{WMT'13 dataset splits.}
	\label{tab:WMT13_splits}
\end{table}

\subsection{IWSLT'16}
\label{sec:transfer}
For inclusion and usage of a general-domain dataset, we adopt the TED talks dataset from the IWSLT 2016
German-English translation task datasets \cite{IWSLT_2016}. We apply the same data processing steps as on the WMT'13 dataset. These steps are simply applying punctuation normalization, tokenization, and data cleaning using the Moses scripts. The dataset statistics can be found in Table \ref{tab:IWSLT16_statistics}.

\begin{table}[t]
\footnotesize
	\centering\begin{tabular}{@{}l|ll@{}}
    \toprule
                                & \multicolumn{2}{c}{Dataset} \\ \midrule
    Language                    & German       & English      \\ \midrule
    No. of Samples              & 196,884      & 196,884      \\
    No. of Samples Preprocessed & 196,884      & 196,884      \\
    No. of Samples Cleaned      & 195,897      & 195,897      \\
    No. of Samples Unique       & 194,226      & 194,226      \\ \bottomrule
    \end{tabular}
	\caption{IWSLT'16 dataset statistics and preprocessing.}
	\label{tab:IWSLT16_statistics}
\end{table}

This processing pipeline results in 194,226 sentence pairs. These pairs are used to provide additional data for the training set of WMT'13.

\paragraph{Why Including the IWSLT'16 Dataset.}
We are motivated by the usage of a general-domain dataset to improve the Transformer model performance on the WMT'13 test set. Our hypothesis is that the general-domain nature of the TED talks data of the IWLST'16 dataset should help the Transformer model learn a better representation of the textual context of both the source and target sentences. This benefits the system in general even if the evaluation is done only using the test set of WMT'13. 
This usage of additional general-domain data is also helpful when the original dataset for the MT task is tiny. This helps reduce the overfitting of the Transformer model to the original dataset.
\section{Experimental Evaluation}

In this section, we present our study on training the Transformer model on the WMT'13 news commentary German-English translation task dataset. We report the effects of including the IWSLT'16 TED talks German-English translation task dataset in training.

\begin{figure*}[t]
    \centering
    \includegraphics[width=0.8\textwidth]{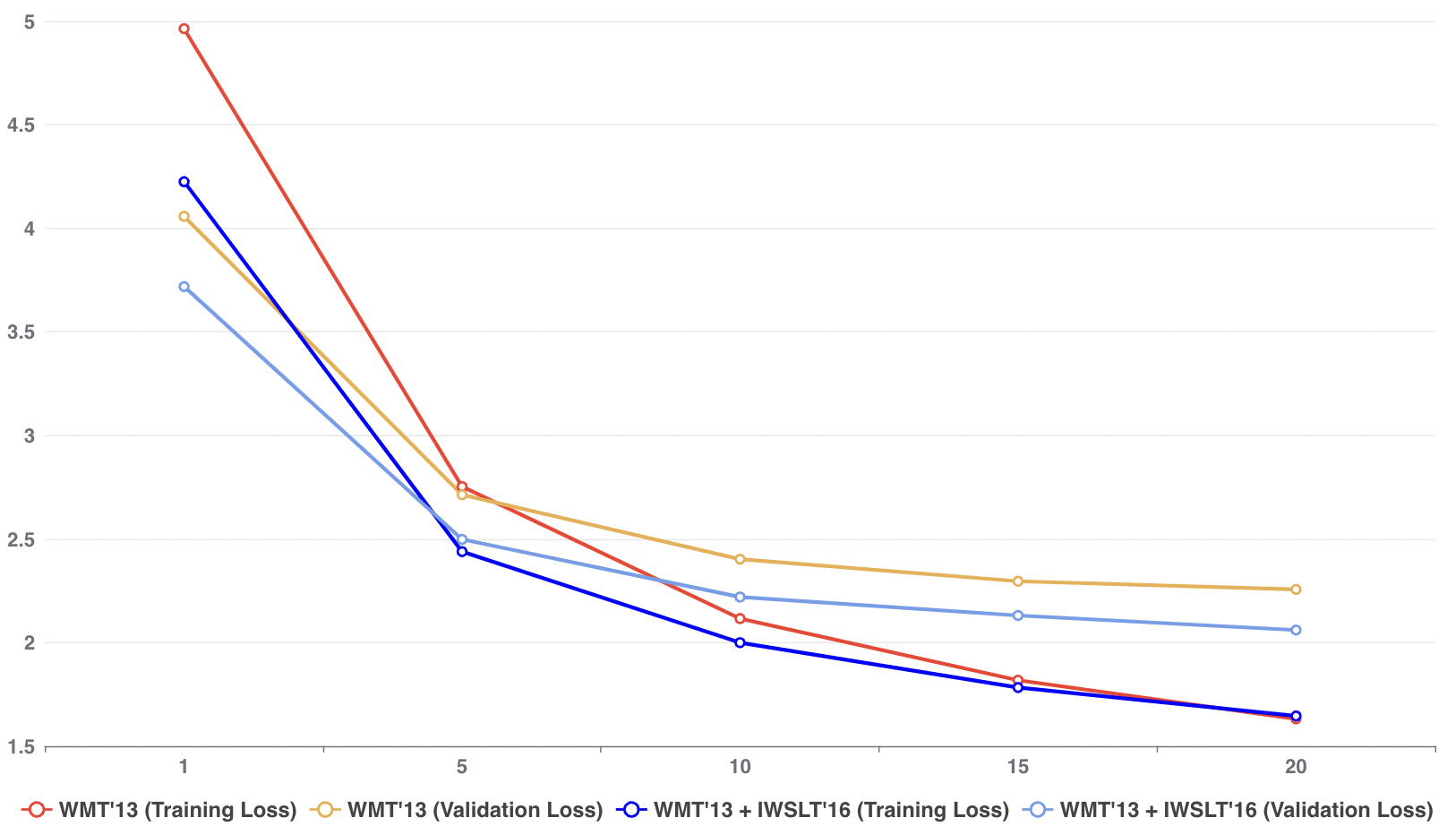}
    \caption{Training and validation losses of the trained Transformer model on WMT'13 vs. WMT'13 plus IWSLT'16.}
    \label{fig:training_validation_losses}
\end{figure*}

\subsection{Experimental Setup}

The settings we specify for the Transformer model can be found in Table \ref{tab:transformer_hyperparameters}. We went previously over most of these parameters in our explanation for the Transformer model. However, the forward expansion factor refers to the number of times to expand the embedding size in the first layer of each point-wise feed-forward network layer inside the Transformer model before collapsing it again to the original embedding size in the second layer.

\begin{table}[t]
\footnotesize
	\centering\begin{tabular}{@{}l|l@{}}
    \toprule
    Hyperparameters                  & Value   \\ \midrule
    Source (German) Vocabulary Size  & 137,485 \\
    Target (English) Vocabulary Size & 56,225  \\
    Embedding Size                   & 512     \\
    No. of Heads                     & 8       \\
    No. of Encoder Layers            & 3       \\
    No. of Decoder Layers            & 3       \\
    Max. Sequence Length             & 100     \\
    Forward Expansion Factor         & 4       \\ \bottomrule
    \end{tabular}
	\caption{Transformer hyperparameters.}
	\label{tab:transformer_hyperparameters}
\end{table}

The hyperparameters used for training are shown in Table \ref{tab:training_hyperparameters}. We do not perform hyperparameter tuning for any of these parameters. We only experiment with modifying the number of epochs as we see later in the results section.

\begin{table}[t]
\footnotesize
	\centering\begin{tabular}{@{}l|l@{}}
    \toprule
    Hyperparameters                  & Value   \\ \midrule
    No. of Epochs                    & 5       \\
    Learning Rate                    & 5e-4    \\
    Batch Size                       & 64      \\
    Dropout                          & 0.1     \\
    Early Stopping                   & True    \\ \bottomrule
    \end{tabular}
	\caption{Training hyperparameters.}
	\label{tab:training_hyperparameters}
\end{table}

\paragraph{Evaluation Metrics.}
The candidate translations produced by the Transformer model are tested using reference translations and assessed by the Bilingual Evaluation Understudy score. In the literature, the BLUE score is widely adopted for comparison across different models of MT. The BLEU score ranges from 0 to 1. We do not want to achieve a score close to 1. The score should be close to the score of a human \cite{QiuMM21}. However, in our case, we focus on comparing the model trained on the WMT'13 dataset vs. the model trained on the WMT'13 plus IWSLT'16 datasets.
When computing the BLEU score, the average of the n-gram precision is computed and presented as $p_{n}$. A word's count in a candidate translation is clipped into the count of the word in its corresponding reference translation. The total count is then divided by the total number of words in the candidate translation. Equation \ref{eq:bleu} shows how the BLEU score is computed, where $N$ is the maximum length of n-grams and $w_{n}$ is the n-th weight with the sum of all the $w_n$ equal to 1.

\begin{equation}
\footnotesize
\begin{split}
    BP = 
    \left\{\begin{matrix}
        1 & if & c > r \\
        e^{1-r/c} & if & c \leq r
    \end{matrix}
    \right.
\end{split}
\label{eq:bp}
\end{equation}

\begin{equation}
\footnotesize
\begin{split}
    BLEU = BP \cdot \exp{(\sum_{n=1}^Nw_{n}\log(p_{n}))}
\end{split}
\label{eq:bleu}
\end{equation}

In the calculation of the BLEU score, the Brevity Penalty (BP) penalizes short candidate translations. The shorter the candidate translation, the higher the penalty. In Equation \ref{eq:bp}, $c$ represents the length of the candidate translation, while $r$ represents the length of the reference translation.

\subsection{Results}

\begin{table}
\footnotesize
	\centering\begin{tabular}{@{}l|l|l@{}}
    \toprule
    Training Dataset  & No. of Epochs & Bleu Score \\ \midrule
    WMT'13            & 5             & 16.1       \\
    WMT'13 + IWSLT'16 & 5             & 18.4       \\
    WMT'13            & 20            & 21.2       \\
    WMT'13 + IWSLT'16 & 20            & 23.1       \\
    WMT'13 + IWSLT'16 & 50            & \textbf{25.8}       \\ \bottomrule
    \end{tabular}
	\caption{BLEU scores on the test set of WMT'13.}
	\label{tab:bleu_scores}
\end{table}

\begin{table*}[t]
\footnotesize
	\centering\begin{tabular}{@{}llll@{}}
    \toprule
    Language              & Training Dataset                                             & Epochs & Sentence                                                                                                                                                                                  \\ \midrule
    German (Source)       & -                                                            & -             & \textbf{\begin{tabular}[c]{@{}l@{}}In letzter Zeit allerdings ist dies schwieriger denn je, ist doch der \\ Goldpreis im letzten Jahrzehnt um über 300 Prozent angestiegen.\end{tabular}} \\ \midrule
    English (Target)      & -                                                            & -             & \textbf{\begin{tabular}[c]{@{}l@{}}Lately, with gold prices up more than 300 \% over the last decade, \\ it is harder than ever.\end{tabular}}                                                                                        \\ \midrule
    English (Translation) & \begin{tabular}[c]{@{}l@{}}WMT'13 + \\ IWSLT'16\end{tabular} & 5             & \begin{tabular}[c]{@{}l@{}}But, over the last few months, this is more difficult, because the more \\ difficult, the net in the last decade has increased by 300 \%.\end{tabular}         \\ \midrule
    English (Translation) & \begin{tabular}[c]{@{}l@{}}WMT'13 + \\ IWSLT'16\end{tabular} & 50            & \begin{tabular}[c]{@{}l@{}}But, more recently, this has become more difficult than ever, but gold \\ price rose more than 300 \% over the past decade.\end{tabular}                       \\ \bottomrule
    \end{tabular}
	\caption{Translation quality progression over the number of epochs.}
	\label{tab:quality_progression}
\end{table*}

\begin{table*}[t]
\footnotesize
	\centering\begin{tabular}{@{}llll@{}}
    \toprule
    Language              & Training Dataset                                             & Epochs & Sentence                                                                                                                                                                                  \\ \midrule
    German (Source)       & -                                                            & -             & \textbf{\begin{tabular}[c]{@{}l@{}}In letzter Zeit allerdings ist dies schwieriger denn je, ist doch der \\ Goldpreis im letzten Jahrzehnt um über 300 Prozent angestiegen.\end{tabular}} \\ \midrule
    English (Target)      & -                                                            & -             & \textbf{\begin{tabular}[c]{@{}l@{}}Lately, with gold prices up more than 300 \% over the last decade, \\ it is harder than ever.\end{tabular}}                                                                                        \\ \midrule
    English (Translation) & WMT'13                                                       & 20             & \begin{tabular}[c]{@{}l@{}}But this is more difficult in recent times, for the price of gold has \\ increased by more than 300 \% over the last decade.\end{tabular}                      \\ \midrule
    English (Translation) & \begin{tabular}[c]{@{}l@{}}WMT'13 + \\ IWSLT'16\end{tabular} & 20             & \begin{tabular}[c]{@{}l@{}}But more recently, this is more difficult than ever, but gold \\  has increased by over 300 \% over the past decade.\end{tabular}                        \\ \bottomrule
    \end{tabular}
	\caption{Qualitative analysis of the trained Transformer model on WMT'13 vs. WMT'13 plus IWSLT'16.}
	\label{tab:qualitative_analysis}
\end{table*}

We first investigate the effect of training the Transformer model on the WMT'13 training set, and then including the IWSLT'16 dataset in the training. We report the results of applying these trained Transfomer models on the test set of the WMT'13 dataset. We adopted the BLEU score in this assessment.

In addition, we experiment with using different numbers of epochs in the training. Table \ref{tab:bleu_scores} compares these models in terms of the BLEU score on the test set of the WMT'13 dataset. Intuitively, the higher the number of epochs, the better the BLEU scores. 

However, for the same number of epochs, the model trained on WMT'13 plus IWLST'16 always outperforms the model trained only on WMT'13 with a gain of around two BLEU score points. We continue the experiments using the best model (trained on WMT'13 plus IWSLT'16) and train it for 50 epochs, achieving a maximum BLEU score result of 25.8.

The results show that training on WMT'13 datset only is not sufficient and that especially including the IWSLT'16 datset helps improve the BLEU scores. This indicates the value of including general-domain dataset in the training data.

This fact is supported by Figure \ref{fig:training_validation_losses}. It shows the progression of the calculated training and validation losses over the number of epochs. The y-axis reflects the cross-entropy loss, and the x-axis reflects the number of epochs. The lines in dark and light red refer to the training and validation losses of the model trained on WMT'13 only, respectively. While the lines in dark and light blue refer to the training and validation losses of the model trained on WMT'13 plus IWSLT'16 datasets, respectively.

If we investigate the validation losses of the two models represented by the light red and blue lines, we find that the model trained on WMT'13 plus IWSLT'16 achieves consistently a lower validation loss than the model trained on WMT'13 only. It always scores about 0.5 lower cross-entropy loss for all the stated number of epochs. In addition, the dark red and blue lines, which show the training losses, demonstrate that the model trained on WMT'13 plus IWSLT'16 trains faster than the model trained on WMT'13 only. It achieves the same loss achieved by the model trained on WMT'13 only in earlier epochs before they both achieve nearly the same loss at the 20th epoch.

\paragraph{Qualitative Analysis.}
In this part, we take a closer look at the translations produced by the Transformer model. We aim to assess how it learns to translate an example sentence and how the quality of the translation improves over the number of epochs, as shown in Table \ref{tab:quality_progression}.

The table shows an example sentence in the source language (German) and its ground truth reference translation in the target language (English). For the model trained for 5 epochs, the resulting translation is not accurate. It misses an important keyword which is "gold". On the other hand, the model trained for 50 epochs can capture this important keyword and the produced sentence overall is more accurate and relevant to the reference translation.

Another assessment is done to compare the Transformer model trained on WMT'13 and the model trained on WMT'13 plus IWSLT'16 for the same number of epochs. The results can be shown in Table \ref{tab:qualitative_analysis}. It shows the same example sentence and reference translation from the previous table.

The results show that the model trained on WMT'13 only gives a translation that misses a relevant keyword "than ever" from the reference sentence, although the overall meaning does not massively change. However, the model trained on WMT'13 plus IWSLT'16 produces a more accurate translation that is more consistent with the reference sentence, and it contains the complete meaning by including the keyword "than ever". This demonstrates the strength of including additional general-domain datasets in the training data to produce more relevant translation sentences.

\section{Conclusion}

In this paper, we presented an explanation for the Transformer model and its components. In particular, we utilized it to obtain a German-English machine translation system. In our experiments, we investigated the effect of including a general-domain dataset in training and whether it helps or harms the performance of the model. The original dataset used to train and evaluate the model comes from the news commentary dataset of the WMT'13 German-English translation task. The additional general-domain dataset deployed is the IWSLT'16 TED talks German-English dataset.
Our results showed that it is helpful to use additional general-domain dataset in the training data. We observe that for the same number of epochs, the model trained with additional general-domain data achieves on average a gain of two BLEU score points on the test set of the WMT'13 dataset. Possible actions of this work can go in the direction of investigating methods to generate synthetic data to augment the general-domain dataset automatically. This should improve and help the Transformer model obtain more relevant translation sentences and generate higher quality rich representations for the textual context of both the source and target sentences, which in return should help with the translation task.
%\input{latex/sections/todo}

%\section*{Acknowledgements}

%\bibliography{anthology,main}
\bibliography{main}

% \input{latex/sections/appendix}

%\appendix
%
%\section{Example Appendix}
%\label{sec:appendix}
%
%This is an appendix.

\end{document}